# Automated detection of motion artifacts in brain MR images using deep learning and explainable artificial intelligence

Marina Manso Jimeno[1,2], Keerthi Sravan Ravi[1,2], Maggie Fung[3], John Thomas Vaughan, Jr.[1,2,4,5], Sairam Geethanath[2,6]*

[1] Department of Biomedical Engineering, Columbia University in the City of New York, 10027, New York, NY, USA

[2] Columbia Magnetic Resonance Research Center, Columbia University in the City of New York, New York, 10027, NY, USA

[3] MR Clinical Solutions, GE Healthcare, New York, NY, USA

[4] Department of Radiology, Columbia University Medical Center, New York, 10032, NY, USA

[5] Zuckerman Institute, Columbia University in the City of New York, New York, 10027, NY, USA

[6] Department of Radiology and Radiological Science, John Hopkins University, Baltimore, 21205, MD, USA

**Corresponding author:**
Sairam Geethanath
sairam.geethanath@jhu.edu
601 N Caroline St,
Baltimore, MD, 21205

# Abstract

Quality assessment, including inspecting the images for artifacts, is a critical step during MRI data acquisition to ensure data quality and downstream analysis or interpretation success. This study demonstrates a deep learning model to detect rigid motion in $T_1$-weighted brain images. We leveraged a 2D CNN for three-class classification and tested it on publicly available retrospective and prospective datasets. Grad-CAM heatmaps enabled the identification of failure modes and provided an interpretation of the model's results. The model achieved average precision and recall metrics of 85% and 80% on six motion-simulated retrospective datasets. Additionally, the model's classifications on the prospective dataset showed a strong inverse correlation (-0.84) compared to average edge strength, an image quality metric indicative of motion. This model is part of the ArtifactID tool, aimed at inline automatic detection of Gibbs ringing, wrap-around, and motion artifacts. This tool automates part of the time-consuming QA process and augments expertise on-site, particularly relevant in low-resource settings where local MR knowledge is scarce.

# Abbreviations

**AES**

Average edge strength

**AUC**

Area under the curve

**AUROC**

Area under the receiver operating characteristic curve

**CNN**

Convolutional neural network

**DL**

Deep learning

**DWI**

Diffusion-weighted imaging

**FOV**

Field of view

**IRB**

Institutional review board

**MNI**

Montreal Neurological Institute

**MR**

Magnetic resonance

**MRI**

Magnetic resonance imaging

**QA**

Quality assessment

**ReLU**

Rectified linear unit

**ROC**

Receiver operating characteristic curve

**ROI**

Region of interest

**TE**

Echo time

**TI**

Inversion time

**TR**

Repetition time

**XAI**

Explainable artificial intelligence

# 1.   Introduction

Quality assessment (QA) of clinical MRI data is essential to avoid any acquisition errors that could compromise the analysis, interpretation, or diagnosis of the images[1–3]. Artifact detection is a critical step of the QA procedure, typically performed during the imaging session while the subject is still inside the scanner in case repetition is required. The number of artifacts and their possible sources make this task time-consuming and highly demanding of locally available MR physics expertise[4,5]. Consequently, it burdens the MR technicians and may result in loss of scan efficiency due to long scanning sessions or patient rescheduling, especially in low-resource settings where personnel lack this expertise or in highly populated regions where high patient throughput is vital[6,7].

Patient voluntary or involuntary motion is a frequent source of artifacts in MR imaging[8], with an estimated annual cost of $115,000 per scanner[9]. Factors such as pain, restlessness, agitation, and anxiety due to long scan times and claustrophobia sensation can contribute to patient movement[10]. Images corrupted by motion may include blurring and ghosting artifacts[8,11,12], degrading image quality, and potentially impacting image interpretation and quantitative analyses[13–15]. The random nature of motion patterns makes their prediction and correction challenging[16]. While prospective and retrospective correction strategies[17,18] can be applied to mitigate motion artifacts, recovering ideal images is not possible once the data is corrupted. Therefore, inline motion detection and control or prevention during image acquisition represent more effective methods to ensure data quality[8,19].

Sequences with long scan durations, such as 3D $T_1$-weighted acquisitions, are particularly susceptible to motion artifacts[20]. Given the key role of this sequence in structural analysis and as anatomical reference images in functional MRI and diffusion-weighted imaging (DWI) studies[21–24], early motion detection is crucial to prevent downstream errors. Automatic QA strategies offer real-time feedback to MR technicians regarding artifact presence and localization, aiding in deciding whether a scan should be repeated. With the advent of deep learning (DL) techniques in medical image processing and analysis, several convolutional neural network (CNN)-based approaches have emerged for automatically identifying rigid motion in brain MRI scans with high classification accuracies[25–29].

Iglesias et al.[25] and Kustner et al.[26] implemented CNNs to derive voxel- and patch-wise motion probability maps, respectively. Fantini et al.[27] and Mohebbian et al.[28] combined the results of four specialized CNNs for ensemble motion classification, while Vakli et al.[29] used a lightweight 3D CNN yielding comparable results to models trained on image quality metrics. Despite the ability of deep CNNs to capture complex data patterns, lightweight model architectures capable of fast inferences are preferred for inline artifact detection during image acquisition[30], especially in low-resource settings where high performance computing systems might be unavailable. This requisite favors volume-wise or slice-wise classification over patch- or voxel-wise approaches, which entail additional layers or processing steps to compute the overall motion score or class[25–27]. With the exception of Mohebbian et al.[28], all methods implemented binary classification to determine data viability, without considering scenarios where motion levels might be tolerable or confined to areas outside the region of interest (ROI). Furthermore, the laborious task of data annotation and the scarcity of motion-corrupted publicly available datasets often led to small prospective datasets[25–27] or the need to synthesize motion artifacts for training, testing, and generalization assessment[28,29]. Thus, to enhance trust in the models' results and facilitate their interpretability[31], two of these methods incorporated explainable artificial intelligence (XAI) tools[26,28].

Our work combines the strengths of previous existing CNN-based methods for detection and localization of motion artifacts in $T_1$-weighted brain images. We leverage a lightweight 2D CNN for three-class classification trained on synthesized motion-corrupted images. We tested our models on motion-free and motion-synthesized images from other publicly available datasets and a prospectively acquired dataset to assess the model's generalization ability to multiple field strengths, vendors, and acquisition parameters. Additionally, we utilize XAI tools to develop trust and interpret the model results. This method advances our ArtifactID tool[32], which includes models for identifying wrap-around and Gibbs ringing artifacts.

## 2. Methods

### 2.1. Datasets

This work utilized both retrospective and prospectively acquired datasets. The retrospective dataset included $T_1$-weighted data from publicly available sources. In particular, we leveraged the IXI[33], HCP[34], ADNI-3[35], SUDMEX[36], SRPBS[37], and LAC[38] datasets, and assume their images are motion-free. We used the first 100 volumes of the IXI dataset and 36 subjects from each of the remaining five public datasets. These contained images from healthy volunteers or an even split of healthy and pathological data, wherever applicable, across three vendors (Siemens, GE, and Philips) and two field strengths (1.5 T and 3 T).

The prospective dataset included artifact-free and artifact-induced data from five healthy volunteers acquired in a GE SIGNA Premier 3 T scanner. Prior to data acquisition, all subjects provided written informed consent in accordance with the protocol approved by the Institutional review board (IRB). The acquisition protocol included three 3D $T_1$-weighted sequences. On the first sequence, the subject was instructed to lay as still as possible to avoid motion artifacts; on the second one, we instructed the subject to slightly move the head during the center portion of the sequence; and on the third one, we encouraged the subject to freely move the head during the entire duration of the sequence. The acquisition parameters were: TE=3.1 ms, TR=7.6 ms, TI=450 ms, matrix size=256x256, number of slices=140.

## 2.2.   Motion synthesis

We followed the motion artifact simulation proposed by Pawar et al.[39] on 3D $T_1$-weighted sagittal images. This method assumes that motion events happen along the phase encoding direction. Involuntary, rigid head motion requires six degrees of freedom, comprising three translations and three rotations[8]. We performed artifact simulation on 100 volumes of the IXI dataset and used 3 motion levels as labels. Labels ranged from 0 (no motion) to 2 (severe motion). It is important to note that motion levels are not strictly equivalent to image quality or clinical acceptance levels. After motion level classification, the user should visually or experimentally assess which artifact level is tolerable for the specific imaging purpose or application.

We generated a gradual motion curve of four motion events for each forward-simulated volume, each event occurring near the k-space center at the 93rd, 118th, 238th, and 163rd phase-encoding lines. For motion class 1, the motion events had random extent of rotation and translation along the three axes of up to 1° and 1 mm, respectively. For motion class 2, the extent of rotation and translation along the three axes was within (3-4)° and (3-4) mm, respectively. We experimentally set these values as they provided the best visual discrimination between motion classes 1 and 2. At each phase encoding step, we applied the corresponding rigid transformation given by the motion curve to the 3D image volume. After each rigid transformation, the relevant k-space lines were concatenated in a new volume that was ultimately converted back to image space to obtain the motion-corrupted volume. Figure 1 illustrates the forward-modeling pipeline.

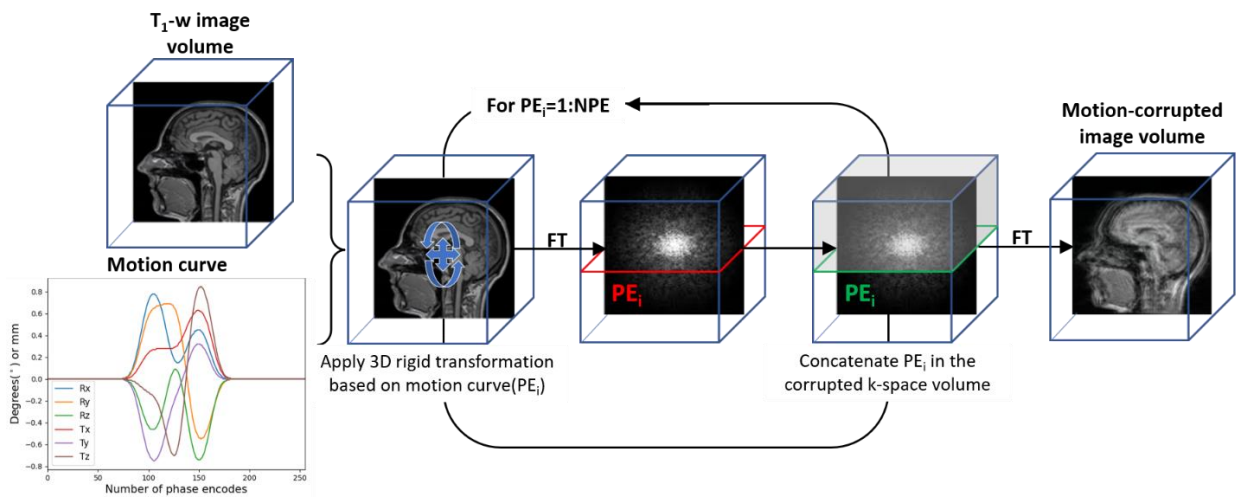

**Figure 1.** Pipeline for motion artifacts forward modeling on sagittal brain $T_1$-weighted images. Based on the gradual motion curve, the 3D volume iteratively undergoes rigid transformation, and the corresponding k-space lines are concatenated in a new motion-corrupted volume.

## 2.3. Model implementation, training, and evaluation

We chose a lightweight 2D CNN model as network architecture for inline multi-class classification of motion artifacts. This is the same model leveraged in previous work[32] for identifying wrap-around and Gibbs ringing artifacts, optimized for fast inference times and deployment in low-resource environments. The model consisted of three convolutional layers with 3x3 kernel size followed by max pooling layers and five fully-connected dense layers with decreasing units. All layers were activated with the rectified linear unit (ReLU)[40] function, and the last dense layer of the model was modified to have three filters to allow a three-class classification. The model, depicted in Figure 1 of ref. 32, was implemented with Keras-Tensorflow 2.1[41] and utilized Adam[42] optimizer to reduce the sparse categorical cross-entropy loss.

We used the IXI dataset for training, validation, and testing. The motion simulation pipeline randomly assigned classes to each subject's volume while ensuring class balance. We split the IXI motion-simulated data subject-wise and maintained class balance within each set. The split proportions were 80-20% between the training and testing sets, and 20% of the training set was used for validation. Before training and testing, each volume was intensity normalized to [0,1]. The total number of slices used for training was 1,116. We trained the model for 20 epochs using a batch size of 32. The number of trainable parameters was 15,481,987.

We evaluated the trained models on the test datasets to obtain the class-wise and overall accuracies. We leveraged Grad-CAM[43] to develop trust in the models' predictions, analyze failure modes, and fine-tune the pre-processing steps. Grad-CAM's heatmaps highlighted regions of the input that most significantly contributed to the model's decision. We obtained these heatmaps for representative cases of correct and incorrect classification. Besides testing the artifact-simulated IXI datasets unseen during training, we tested the models on other artifact-free and artifact-corrupted datasets to assess their performance on out-of-distribution data. Out-of-distribution testing of the motion models included testing on clean and motion-simulated publicly available datasets and motion-induced prospectively acquired datasets.

## 2.4.  Assessment of classification performance

Multiple metrics were considered to evaluate the model's classification performance depending on the testing dataset. Besides calculating the training and validation accuracies, we obtained the macro-averaged precision and recall metrics for the motion-synthesized public data testing sets. Additionally, for the simulated IXI test set, we reported the area under the receiver operating characteristic curve (AUROC).

Due to the lack of annotations by an image quality specialist for the prospective dataset, we compared the models' predictions to the image quality metric average edge strength (AES). This metric is an indirect metric of blurriness in the image and has been reported to correlate with motion[13,44]. To assess the agreement between AES as a motion metric and our model's predictions, we computed the non-parametric Spearman's rank correlation coefficient and Cohen's Kappa statistic between the two. We modified AES to be a categorical variable by thresholding the histogram using Otsu's method and binning the values into the three classes for the calculation of the Cohen's Kappa values.

## 3. Results

The model achieved training and validation accuracies of 100% and 96.5%, respectively. Macro-averaged precision and recall metrics on the IXI motion-simulated test set were 98.29% and 98.33%. Figure 2 shows the receiver operating characteristic (ROC) curve and the confusion matrix resulting from inference on the IXI forward-modeled test set. The area under the curve (AUC) scores were 1.00 for the no-motion class and 0.98 for motion classes 1 and 2.

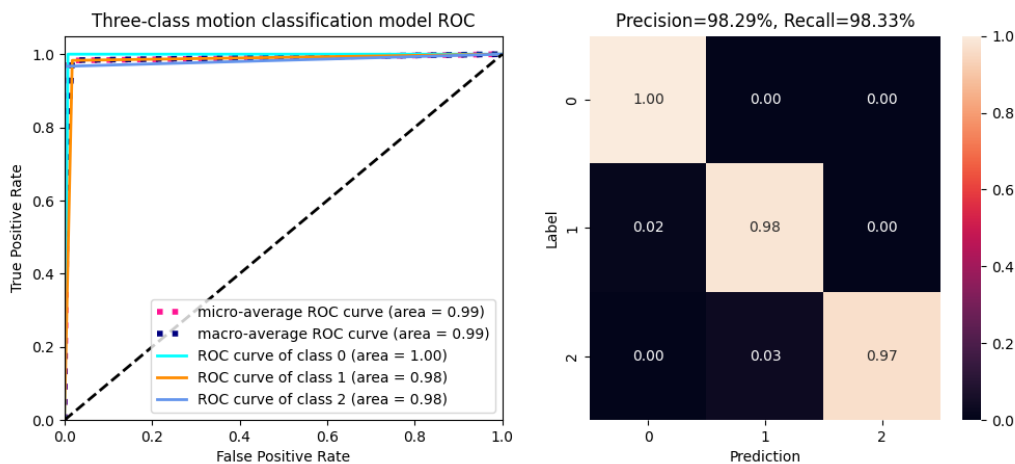

**Figure 2.** Receiver operating characteristic (ROC) curve and confusion matrix for the three-class motion artifact classification model on the forward-modeled IXI test set.

Figure 3 illustrates the Grad-CAM visual explanations after inference on the IXI motion-simulated test set. A small fraction of the no-motion and motion class 1 slices were incorrectly classified as motion class 1 (Figure 3ii) and motion class 2 (Figure 3iv), respectively. However, no slices of classes 1 or 2 were classified as class 0 (motion-free). The Grad-CAM heatmaps indicate that the model pays attention to regions within the anatomy when motion artifacts are identified, whereas background regions are highlighted in the heatmaps of slices classified as no-motion. We observed that the magnitude of the Grad-CAM heatmaps increases from class 0 to 2 as the severity of the artifact grows.

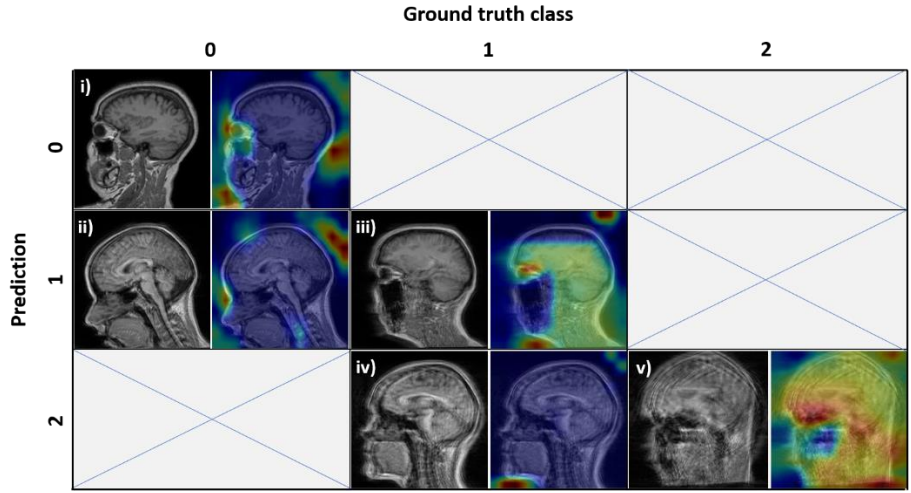

**Figure 3.** Representative Grad-CAM heatmaps highlighting two levels of motion artifacts on the IXI motion-simulated test sets. The heatmaps are overlaid on input slices and displayed next to the original grayscale images. Crossed cells mean there were no misclassified samples.

As expected, testing the model on the other five forward-modeled public datasets resulted in a reduction in performance (Table I). Inference on the ADNI-3 dataset resulted in the best performance after inference on the IXI dataset, with precision and recall metrics above 90%. In contrast, average precision and recall values for the HCP, LAC, SUDMEX, and SRPBS datasets were 81.3% and 71.3%, respectively.

**Table 1.** Precision and recall metrics after inference on the motion-simulated retrospective test sets.

| Dataset | Precision | Recall |
|---------|-----------|--------|
| IXI | 0.983 | 0.983 |
| ADNI-3 | 0.940 | 0.927 |
| HCP | 0.825 | 0.752 |
| LAC | 0.769 | 0.661 |
| SUDMEX | 0.762 | 0.693 |
| SRPBS | 0.825 | 0.725 |

Motion classification performance initially dropped after testing on the prospectively acquired dataset. After inference, we observed good accuracy for the data acquired with the first sequence (assumed motion-free), with at least 95% of the total slices classified as class 0 for all subjects. However, large percentages of slices from sequence 2 and sequence 3 were classified as motion-free but were visibly corrupted by motion. Figure 4 shows the distribution of the reference motion estimation metric AES for the slices classified as class 0, 1, and 2, along with the confusion matrix comparing the model and AES classifications. We obtained a Spearman correlation coefficient of -0.58 (p-value < 0.01), indicating a moderate inverse association between the two variables. The figure shows a large percentage of slices with AES values typical of motion classes 1 or 2 being incorrectly classified as no motion, resulting in poor agreement (K=0.16) between the two classification methods.

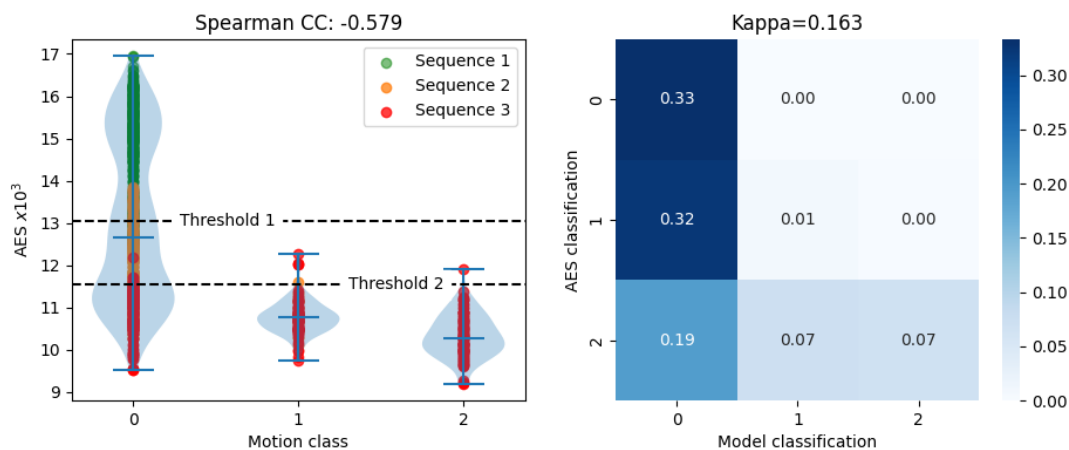

**Figure 4.** (Left) Violin plots of the average edge strength (AES) distribution for slices of the prospective dataset classified as class 0, class 1, and class 2. The dotted lines represent the computed thresholds for AES classification. (Right) Confusion matrix of the model versus AES classification.

Evaluating the Grad-CAM heatmaps of misclassified instances (Figure 5 ii)-iii)) suggested that the model's classification is based on the background of the images rather than on the head or brain. We noticed that the training data was characterized by smaller field of views (FOV), well centered, and more fitted to the subject's head with reduced background regions on each slice. To assess if FOV was the confounding factor, we manually cropped the background while maintaining squared matrix sizes and repeated the analysis.

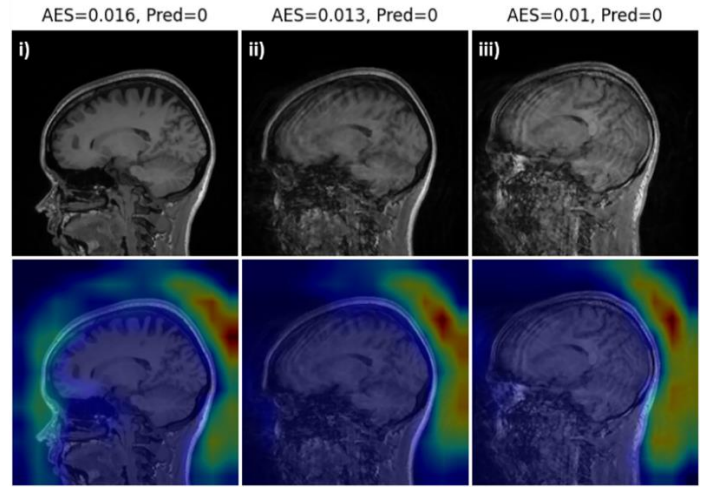

**Figure 5.** Representative slices from the prospective dataset and their respective Grad-CAM heatmaps. Average edge strength (AES) values for cases ii) and iii) indicate motion presence, but the heatmaps highlight the background, suggesting no artifact was detected by the model.

After cropping, the Spearman correlation coefficient obtained a Spearman correlation coefficient increased to -0.84 (p-value < 0.01), indicating a very strong inverse association between the model's predictions and AES. Figure 6 portrays the improved correlation between motion classes and AES, which decreases for slices with mild and severe motion from sequences 1 and 2. We obtained a Kappa value of 0.415, indicating moderate agreement between AES and the model's classifications. The two methods agree in classifying motion-free versus motion-corrupted slices but exhibit differences in their discrimination between mild and severe motion.

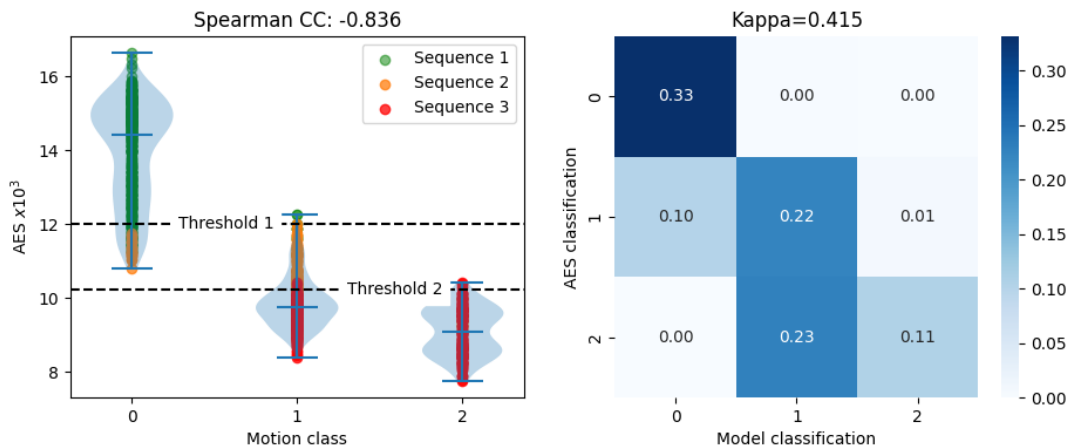

**Figure 6.** (Left) Violin plots of the Average edge strength (AES) distribution for slices of the prospective dataset after cropping and FOV centering classified as class 0, class 1, and class 2. The dotted lines represent the thresholds for AES classification. (Right) Confusion matrix of the model versus AES classification.

After these modifications, the Grad-CAM heatmaps (Figure 7) showed an increase in heatmap magnitude with motion severity, highlighting areas of motion artifact within the anatomy for slices classified as classes 1 and 2.

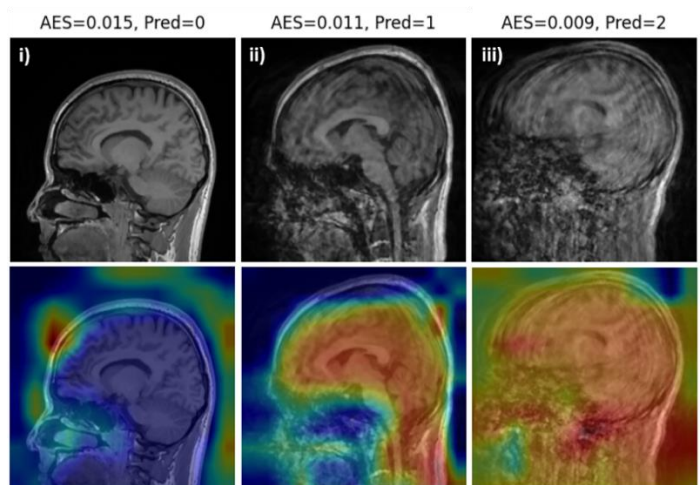

**Figure 7.** Representative slices of the prospective dataset and their respective Grad-CAM heatmaps after fitting and centering the anatomy within the FOV. Average edge strength (AES) value ranges present improved agreement with the model's predictions.

## 4.    Discussion

This work demonstrates a DL approach for detecting motion artifacts in $T_1$-weighted brain MR images. This model builds on previous work for ArtifactID, a tool for inline identification and localization of artifacts, including wrap-around and Gibbs ringing. We trained a simple, lightweight 2D CNN to assess the presence of three classes of motion artifacts slice-wise and leveraged Grad-CAM heatmaps to localize the artifacts and interpret the model's classification.

We synthesized motion on the publicly available IXI dataset to train and test the model. Inference on the IXI test set resulted in optimal classification performance (Figure 2), validated by Grad-CAM heatmaps of each class, which show increased magnitude with increased artifact severity (Figure 3). Testing the model on other motion-simulated retrospective datasets led to a modest drop in performance compared to the IXI test set. Still, it achieved average precision and recall metrics of 82.42% and 75.16%, respectively. To further assess the generalization ability of the method, we performed inference on a prospective dataset and observed a strong inverse association between the model's classifications and AES.

Our model's classification performance is comparable to other methods employing more sophisticated architectures or heavier models[27,28]. We adopted slice-wise detection and tailored the network's architecture for inline motion detection and optimal efficiency during the scanning session. Patch-wise training allows the localization of the artifact within the slice and ensures that input dimensions remain independent of the acquisition matrix size[26]. However, this approach results in longer inference times due to the increased number of samples and requires additional post-processing steps to compute an overall motion score and reassemble the patches into the slice or volume for interpretation. In contrast, our model achieves rapid processing times, averaging 1.8 seconds per volume for the prospective dataset on an Intel Xeon ES-2690 v4 CPU with one NVIDIA Quadro M6000 24 GB GPU. Unlike existing methods primarily focused on binary classification[25–27,29], our model discriminates between motion-free data and the presence of mild and severe artifacts. This capability facilitates a more flexible QA of the images based on the scan purpose and ROI. Furthermore, while other methods are often constrained by small or imbalanced prospective datasets[26,27,29] or rely uniquely on synthesized data cohorts[28], our utilization of Grad-CAM heatmaps and extensive testing across both retrospective and prospective datasets allow the assessment of the method's robustness and its ability to generalize across diverse acquisition parameters, system vendors, field strengths, and clinical conditions.

The performance decline observed in four motion-synthesized retrospective datasets (HCP, SUDMEX, LAC, and SRPBS) could stem from differences in data distribution compared to the IXI training dataset. In these datasets, subjects' faces are masked, and some are registered to other image spaces, such as the Montreal neurological institute (MNI) space. To test this hypothesis, we conducted two additional experiments: first, we performed inference on the data, assumed motion-free, and calculated the percentage of total slices for each class; second, we defaced the ADNI-3 dataset and repeated the first experiment. Figure 8 displays the results of the first experiment. Pre-processed datasets exhibit higher percentages of slices classified as motion-corrupted (class 1 and 2) than the IXI and ADNI-3 datasets. Images previously co-registered show background pixels around the anatomy, including at the base of the slice where the neck should naturally continue (yellow ellipses in Figure 8). The model tends to misclassify these slices as severely motion-corrupted since some instances of motion-synthesized volumes with large ranges of translation in the training set display the same effect of detachment from the image base (Figure 3v)). Additionally, we presume that smoothing steps during defacing or co-registration may induce blurring, which the model may interpret as mild motion. These insights are validated by our second experiment, for which inference on the defaced ADNI-3 dataset led to a notable performance drop, with 69% of total slices classified as class 0 and 13% and 18% as classes 1 and 2, respectively.

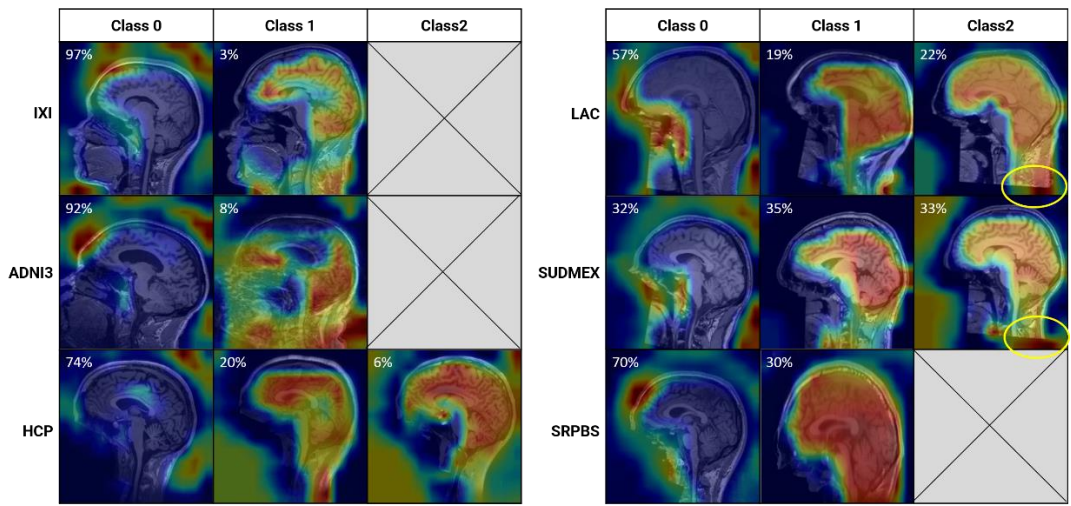

**Figure 8.** Representative Grad-CAM heatmaps after inference on the assumed artifact-free retrospective test sets. The numbers on the top-left corner of the images correspond to the percentage of slices classified as the corresponding class. The yellow ellipses point to areas highlighted by the heatmaps characteristic of transformations during pre-processing steps.

The motion simulation pipeline was optimized to replicate artifacts induced by subject motion in 3D acquisitions. The type (gradual versus abrupt), extent, and timing of the motion event along the encoding of the k-space trajectory all influence the appearance and severity of the artifact, posing challenges to the labeling of the synthetic images and sometimes resulting in inaccuracies. Despite efforts to mitigate the effect of these confounding factors by confining the motion events to specific phase-encoding lines and constraining the extent of rotation and translation per class, this remains a limitation of the study and could contribute to disagreements between the model's prediction and AES values in the prospective set.

Evaluating the model across various publicly available datasets and the prospective dataset revealed its susceptibility to image pre-processing steps. Sensitivity was notably observed towards variations in background/anatomy proportion within the slices, facial features masking, and transformations resulting from co-registration. Testing on a variety of distributions and leveraging Grad-CAM heatmaps helped identify these sensitivities and refine data pre-processing. Given our focus on artifact identification from raw DICOM images immediately post-acquisition, we perceive this challenge as manageable. Nevertheless, future model iterations can address this issue by training the model on a more heterogeneous dataset or implementing data augmentation techniques.

The lack of expert data annotations for the prospective dataset posed challenges in evaluating classification performance. To address this limitation, we opted to employ a heuristic metric to estimate motion in the images. We chose AES due to the simplicity of its implementation and its ability to quantify blurring, a characteristic of motion-corrupted images. While statistical analysis demonstrated agreement between model classifications and AES, qualitative examination of slices classified as class 1 and 2 revealed inconsistencies between AES measurements and motion severity (Figure 9 i)-ii), Figure 9 iii)-iv)). Thus, while AES serves as a helpful reference in the absence of annotations, it cannot be deemed as ground truth. Additionally, standardizing motion severity levels across subjects proved challenging in this proof-of-concept prospective study. Automatic system post-processing steps such as signal averaging could also impact motion severity before exporting the data for inference. Future prospective studies will assess the model's performance after deployment in a clinical setting, including quantification of its impact to scan efficiency and comparison of the model's results to image quality specialist's annotations.

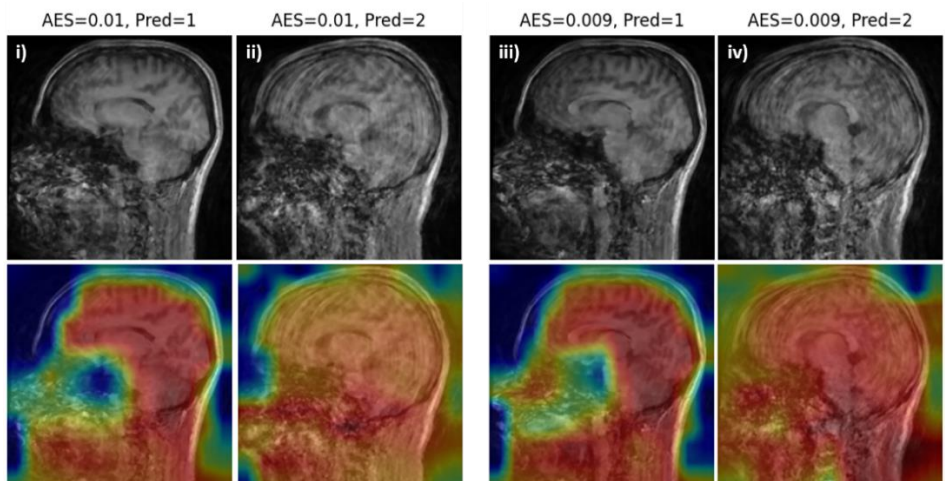

**Figure 9.** Representative Grad-CAM heatmaps after inference on the prospective dataset. Slices with the same average edge strength (AES) motion score present visually diverse severity, correctly captured by the model prediction.

Our work presents a simple, multi-class classification model for identifying motion artifacts in brain MR images. The model features are tailored for seamless integration into the clinical workflow, aiming to identify artifacts as part of the QA task following each sequence acquisition. Deploying this and other ArtifactID models could enhance scan efficiency by flagging artifact-corrupted images while the patient is still inside the scanner, allowing timely reacquisition. Additionally, this tool can augment expertise in low-resource settings by assisting less experienced MR technicians in performing this QA task.

## 5.    Conclusion

The DL model proposed in this work, part of ArtifactID, demonstrated the ability to detect three levels of motion artifacts in $T_1$-weighted brain MR images, achieving high accuracies during training and testing on various motion-simulated retrospective datasets. Our model demonstrates strong potential for generalization to diverse sequence parameters, vendors, and field strengths, and the use of Grad-CAM allowed early identification of failure modes and the fine-tuning of pre-processing requirements. Additionally, its classification performance on the prospective dataset exhibited agreement with AES, a heuristic motion estimation metric. By providing a simple yet effective tool for artifact identification, our model aims to streamline the quality assessment process in MRI scanning sessions, ultimately improving scan efficiency and augmenting expertise in clinical settings.

# References

1.  Sinha N, Ramakrishnan AG. Quality assessment in magnetic resonance images. *Crit Rev Biomed Eng*. 2010;38(2):127-141. doi:10.1615/CritRevBiomedEng.v38.i2.20

2.  Obuchowicz R, Oszust M, Bielecka M, Bielecki A, Piórkowski A. Magnetic resonance image quality assessment by using non-maximum suppression and entropy analysis. *Entropy*. 2020;22(2). doi:10.3390/e22020220

3.  Esteban O, Birman D, Schaer M, Koyejo OO, Poldrack RA, Gorgolewski KJ. MRIQC: Advancing the automatic prediction of image quality in MRI from unseen sites. *PLoS One*. 2017;12(9):e0184661. doi:10.1371/journal.pone.0184661

4.  Lei K, Syed AB, Zhu X, Pauly JM, Vasanawala SS. Artifact- and content-specific quality assessment for MRI with image rulers. *Med Image Anal*. 2022;77:102344. doi:10.1016/j.media.2021.102344

5.  Stadler A, Schima W, Ba-Ssalamah A, Kettenbach J, Eisenhuber E. Artifacts in body MR imaging: Their appearance and how to eliminate them. *Eur Radiol*. 2007;17(5):1242-1255. doi:10.1007/s00330-006-0470-4

6.  Geethanath S, Vaughan JT. Accessible magnetic resonance imaging: A review. *J Magn Reson Imaging*. 2019;49(7):65-77. doi:10.1002/jmri.26638

7.  Manso Jimeno M, Vaughan JT, Geethanath S. Superconducting magnet designs and MRI accessibility: A review. *NMR Biomed*. 2023;36(9):e4921. doi:10.1002/nbm.4921

8.  Zaitsev M, Maclaren J, Herbst M. Motion artifacts in MRI: A complex problem with many partial solutions. *J Magn Reson Imaging*. 2015;42(4):887-901. doi:10.1002/jmri.24850

9.  Andre JB, Bresnahan BW, Mossa-Basha M, et al. Toward quantifying the prevalence, severity, and cost associated with patient motion during clinical MR examinations. *J Am Coll Radiol*. 2015;12(7):689-695. doi:10.1016/j.jacr.2015.03.007

10. Munn Z, Jordan Z. The patient experience of high technology medical imaging: a systematic review of the qualitative evidence. *JBI Database Syst Rev Implement Reports*. 2011;9(19):631-678. doi:10.11124/01938924-201109190-00001

11. Wood ML, Henkelman RM. MR image artifacts from periodic motion. *Med Phys*. 1985;12(2):143-151. doi:10.1118/1.595782

12. Van de Walle R, Lemahieu I, Achten E. Magnetic resonance imaging and the reduction of motion artifacts: review of the  principles. *Technol Heal care  Off J Eur Soc  Eng Med*. 1997;5(6):419-435.

13. Pollak C, Kügler D, Breteler MMB, Reuter M. Quantifying MR head motion in the Rhineland Study – A robust method for population cohorts. *Neuroimage*. 2023;275:120176. doi:10.1016/j.neuroimage.2023.120176

14. Havsteen I, Ohlhues A, Madsen KH, Nybing JD, Christensen H, Christensen A. Are movement artifacts in magnetic resonance imaging a real problem?-a narrative review. *Front Neurol*. 2017;8(MAY):232. doi:10.3389/fneur.2017.00232

15. Reuter M, Tisdall MD, Qureshi A, Buckner RL, van der Kouwe AJW, Fischl B. Head motion during MRI acquisition reduces gray matter volume and thickness estimates. *Neuroimage*. 2015;107:107-115. doi:10.1016/j.neuroimage.2014.12.006

16. Chang Y, Li Z, Saju G, Mao H, Liu T. Deep learning-based rigid motion correction for magnetic resonance imaging: A survey. *Meta-Radiology*. 2023;1(1):100001. doi:10.1016/j.metrad.2023.100001

17. Godenschweger F, Kägebein U, Stucht D, et al. Motion correction in MRI of the brain. *Phys Med Biol*. 2016;61(5):R32-R56. doi:10.1088/0031-9155/61/5/R32

18. Maclaren J, Herbst M, Speck O, Zaitsev M. Prospective motion correction in brain imaging: A review. *Magn Reson Med*. 2013;69(3):621-636. doi:10.1002/mrm.24314

19. Haskell MW. Retrospective motion correction for magnetic resonance imaging. 2019.

20. Ai L, Craddock RC, Tottenham N, et al. Is it time to switch your T1W sequence? Assessing the impact of prospective motion correction on the reliability and quality of structural imaging. *Neuroimage*. 2021;226:117585. doi:https://doi.org/10.1016/j.neuroimage.2020.117585

21. Glasser MF, Sotiropoulos SN, Wilson JA, et al. The minimal preprocessing pipelines for the Human Connectome Project. *Neuroimage*. 2013;80:105-124. doi:10.1016/j.neuroimage.2013.04.127

22. Van Essen DC, Ugurbil K, Auerbach E, et al. The Human Connectome Project: A data acquisition perspective. *Neuroimage*. 2012;62(4):2222-2231. doi:10.1016/j.neuroimage.2012.02.018

23. Rykhlevskaia E, Gratton G, Fabiani M. Combining structural and functional neuroimaging data for studying brain connectivity: A review. *Psychophysiology*. 2008;45(2):173-187. doi:10.1111/j.1469-8986.2007.00621.x

24. Bassett DS, Bullmore ET. Human brain networks in health and disease. *Curr Opin Neurol*. 2009;22(4):340-347. doi:10.1097/WCO.0b013e32832d93dd

25. Iglesias JE, Lerma-Usabiaga G, Garcia-Peraza-Herrera LC, Martinez S, Paz-Alonso PM. Retrospective head motion estimation in structural brain MRI with 3D CNNs. In: *Lecture Notes in Computer Science (Including Subseries Lecture Notes in Artificial Intelligence and Lecture Notes in Bioinformatics)*. Vol 10434 LNCS. ; 2017:314-322. doi:10.1007/978-3-319-66185-8_36

26. Küstner T, Liebgott A, Mauch L, et al. Automated reference-free detection of motion artifacts in magnetic resonance images. *Magn Reson Mater Physics, Biol Med*. 2018;31(2):243-256. doi:10.1007/s10334-017-0650-z

27.    Fantini I, Rittner L, Yasuda C, Lotufo R. Automatic detection of motion artifacts on MRI using Deep CNN. *2018 Int Work Pattern Recognit Neuroimaging, PRNI 2018*. 2018:1-4. doi:10.1109/PRNI.2018.8423948

28.    Mohebbian MR, Walia E, Habibullah M, Stapleton S, Wahid KA. Classifying MRI motion severity using a stacked ensemble approach. *Magn Reson Imaging*. 2021;75:107-115. doi:10.1016/j.mri.2020.10.007

29.    Vakli P, Weiss B, Szalma J, et al. Automatic brain MRI motion artifact detection based on end-to-end deep learning is similarly effective as traditional machine learning trained on image quality metrics. *Med Image Anal*. 2023;88:102850. doi:10.1016/j.media.2023.102850

30.    Li J, Zheng J, Ding M, Yu H. Multi-branch sharing network for real-time 3D brain tumor segmentation. *J Real-Time Image Process*. 2021;18(4):1409-1419. doi:10.1007/s11554-020-01049-9

31.    Aggarwal K, Manso Jimeno M, Ravi KS, Gonzalez G, Geethanath S. Developing and deploying deep learning models in brain magnetic resonance imaging: A review. *NMR Biomed*. 2023;36(12):e5014. doi:https://doi.org/10.1002/nbm.5014

32.    Manso Jimeno M, Ravi KS, Jin Z, Oyekunle D, Ogbole G, Geethanath S. ArtifactID: Identifying artifacts in low-field MRI of the brain using deep learning. *Magn Reson Imaging*. 2022;89. doi:10.1016/j.mri.2022.02.002

33.    IXI dataset. https://brain-development.org/ixi-dataset/.

34.    Fan Q, Witzel T, Nummenmaa A, et al. MGH-USC Human Connectome Project datasets with ultra-high b-value diffusion MRI. *Neuroimage*. 2016;124(Pt B):1108-1114. doi:10.1016/j.neuroimage.2015.08.075

35.    Petersen RC, Aisen PS, Beckett LA, et al. Alzheimer's Disease Neuroimaging Initiative (ADNI): Clinical characterization. *Neurology*. 2010;74(3):201-209. doi:10.1212/WNL.0b013e3181cb3e25

36.    Angeles-Valdez D, Rasgado-Toledo J, Issa-Garcia V, et al. The Mexican magnetic resonance imaging dataset of patients with cocaine use disorder: SUDMEX CONN. *Sci Data*. 2022;9(1):133. doi:10.1038/s41597-022-01251-3

37.    Tanaka SC, Yamashita A, Yahata N, et al. A multi-site, multi-disorder resting-state magnetic resonance image database. *Sci Data*. 2021;8(1):227. doi:10.1038/s41597-021-01004-8

38.    Di Noto T, Marie G, Tourbier S, et al. Towards Automated Brain Aneurysm Detection in TOF-MRA: Open Data, Weak Labels, and Anatomical Knowledge. *Neuroinformatics*. 2023;21(1):21-34. doi:10.1007/s12021-022-09597-0

39.    Pawar K, Chen Z, Shah NJ, Egan GF. Suppressing motion artefacts in MRI using an Inception-ResNet network with motion simulation augmentation. *NMR Biomed*.

2022;35(4):e4225. doi:10.1002/nbm.4225

40.    Agarap AF. Deep learning using rectified linear units (ReLU). *arXiv*. 2018.

41.    Abadi M, Agarwal A, Barham P, et al. TensorFlow: Large-Scale Machine Learning on Heterogeneous Distributed Systems. *arXiv*. 2016. http://arxiv.org/abs/1603.04467.

42.    Kingma DP, Ba J. Adam: A method for stochastic optimization. *arXiv*. 2014.

43.    Selvaraju RR, Cogswell M, Das A, Vedantam R, Parikh D, Batra D. Grad-CAM: Visual Explanations from Deep Networks via Gradient-Based Localization. In: *Proceedings of the IEEE International Conference on Computer Vision*. Vol 2017-Octob. ; 2017:618-626. doi:10.1109/ICCV.2017.74

44.    Zacà D, Hasson U, Minati L, Jovicich J. Method for retrospective estimation of natural head movement during structural MRI. *J Magn Reson Imaging*. 2018;48(4):927-937. doi:10.1002/jmri.25959